\documentclass{article} 
\usepackage{iclr2017_conference,times}
\usepackage{hyperref}
\usepackage{url}
\usepackage{amsfonts}
\usepackage{amsmath}
\usepackage[]{algorithm2e}
\usepackage{tabularx}
\usepackage{ragged2e}  
\newcolumntype{Y}{>{\RaggedRight\arraybackslash}X} 
\usepackage{booktabs}  
\usepackage{soul}
\usepackage{color}
\usepackage{xcolor}
\usepackage{comment}
\DeclareRobustCommand{\hlcyan}[1]{{\sethlcolor{cyan}\hl{#1}}}
\newcommand{\topbottom}[2]{\vtop{\hbox{\strut #1}\hbox{\strut #2}}}
\iclrfinalcopy

\title{Automatic Rule Extraction from Long Short Term Memory Networks}

\author{W. James Murdoch \thanks{Work started during an internship at Facebook AI Research} \\
Department of Statistics\\
UC Berkeley\\
Berkeley, CA 94709, USA \\
jmurdoch@berkeley.edu 
\And
Arthur Szlam \\
Facebook AI Research \\
New York City, NY, 10003 \\
aszlam@fb.com\\
}

\begin{document}

\maketitle

\begin{abstract}
Although deep learning models have proven effective at solving problems in natural language processing, the mechanism by which they come to their conclusions is often unclear.   As a result, these models are generally treated as black boxes, yielding no insight of the underlying learned patterns.  In this paper we consider Long Short Term Memory networks (LSTMs) and demonstrate a new approach for tracking the importance of a given input to the LSTM for a given output. By identifying consistently important patterns of words, we are able to distill state of the art LSTMs on sentiment analysis and question answering into a set of representative phrases. This representation is then quantitatively validated by using the extracted phrases to construct a simple, rule-based classifier which approximates the output of the LSTM.
\end{abstract}

\section{Introduction}
Neural network language models, especially recurrent neural networks (RNN), are now standard tools for natural language processing.  Amongst other things, they are used for translation \cite{seq2seq}, language modelling \cite{languageModels}, and question answering \cite{wikireading}.  In particular, the Long Short Term Memory (LSTM) \cite{lstm} architecture has become a basic building block of neural NLP.  Although LSTM's are often the core of state of the art systems, their operation is not well understood.  
Besides the basic desire from a scientific viewpoint to clarify their workings, it is often the case that it is important to understand why a machine learning algorithm made a particular choice.  Moreover, LSTM's can be costly to run in production compared to discrete models with lookup tables and pattern matching.  

In this work, we describe a novel method for  visualizing the importance of specific inputs for determining the output of an LSTM. We then demonstrate that, by searching for phrases which are consistently important, the importance scores can be used to extract simple phrase patterns consisting of one to five words from a trained LSTM. The phrase extraction is first done in a general document classification framework on two different sentiment analysis datasets. We then demonstrate that it can also be specialized to more complex models by applying it to WikiMovies, a recently introduced question answer dataset. To concretely validate the extracted patterns, we use them as input to a rules-based classifier which approximates the performance of the original LSTM.

\section{Related Work}
\label{sec:relatedWork}
There are two lines of related work on visualizing LSTMs. First, \cite{Rush} and \cite{Karpathy} analyse the movement of the raw gate activations over a sequence. \cite{Karpathy} is able to identify co-ordinates of $c_t$ that correspond to semantically meaningful attributes such as whether the text is in quotes and how far along the sentence a word is. However, most of the cell co-ordinates are harder to interpret, and in particular, it is often not obvious from their activations which inputs are important for specific outputs. 

Another approach that has emerged in the literature \cite{Rei} \cite{Denil} \cite{McCallum} is for each word in the document, looking at the norm of the derivative of the loss function with respect to the embedding parameters for that word. This bridges the gap between high-dimensional cell state and low-dimensional outputs.  These techniques are general- they are applicable to visualizing the importance of sets of input coordinates to output coordinates of any differentiable function.  In this work, we  describe techniques that are designed around the structure of LSTM's, and show that they can give better results in that setting.

A recent line of work \cite{baidu} \cite{wikireading} \cite{squad} \cite{kvMem} has focused on neural network techniques for extracting answers directly from documents. Previous work had focused on Knowledge Bases (KBs), and techniques to map questions to logical forms suitable for querying them. Although they are effective within their domain, KBs are inevitably incomplete, and are thus an unsatisfactory solution to the general problem of question-answering. Wikipedia, in contrast, has enough information to answer a far broader array of questions, but is not as easy to query. Originally introduced in \cite{kvMem}, the WikiMovies dataset consists of questions about movies paired with Wikipedia articles.

\section{Word Importance Scores in LSTMs}
We present a novel decomposition of the output of an LSTM into a product of factors, where each term in the product can be interpreted as the contribution of a particular word. Thus, we can assign importance scores to words according to their contribution to the LSTM's prediction

\subsection{Long Short Term Memory Networks}

Over the past few years, LSTMs have become a core component of neural NLP systems. Given a sequence of word embeddings 
$x_1,...,x_T \in \mathbb{R}^d$, an LSTM processes one word at a time, keeping track of cell and state vectors $(c_1, h_1),...,(c_T, h_T)$ 
which contain information in the sentence up to word $i$. $h_t$ and  $c_t$ are computed as a function of $x_t, c_{t - 1}$ using the below updates
\begin{align}
f_t & = \sigma(W_f x_t + V_f h_{t - 1} + b_f) \\
i_t & = \sigma(W_i x_t + V_i h_{t - 1} + b_i) \\
o_t & = \sigma(W_o x_t + V_o h_{t - 1} + b_o) \\
\tilde{c}_t & = \tanh(W_c x_t + V_c h_{t - 1} + b_c) \\
c_t & = f_t c_{t - 1} + i_t \tilde{c}_t \\
h_t & = o_t \odot \tanh(c_t)
\end{align}

As initial values, we define $c_0=h_0=0$. After processing the full sequence, a probability distribution over $C$ classes is specified by $p$, with 
\begin{equation}
p_i = \text{SoftMax}(W h_T) = \frac{e^{W_i h_T}}{\sum_{j = 1} ^ C e^{W_j h_t}} \label{eq:output_prob}
\end{equation}
where $W_i$ is the $i$'th row of the matrix $W$

\subsection{Decomposing the Output of a LSTM}
\label{sec:decomposing}
We now show that we can decompose the numerator of $p_i$ in Equation  \ref{eq:output_prob} into a product of factors, and interpret those factors as the contribution of individual words to the predicted probability of class $i$. Define 
\begin{equation}\label{eq:beta} 
\beta_{i,j} = \exp\left(W_i (o_T \odot (\tanh(c_j)- \tanh(c_{j-1}))\right),
\end{equation}
 so that 
\[\exp( W_ih_T) = \exp\left(\sum_{j=1}^T W_i (o_T \odot (\tanh(c_j) - \tanh(c_{j-1}))\right) = \prod_{j=1}^T \beta_{i,j}.\]

 As $\tanh(c_j) - \tanh(c_{j-1})$ can be viewed as the update resulting from word $j$, so $\beta_{i,j}$ can be interpreted as the multiplicative contribution to $p_{i}$ by word $j$. 
\subsection{An Additive Decomposition of the LSTM cell}
\label{sec:additive}
We will show below that the $\beta_{i,j}$  capture some notion of the importance of a word to the LSTM's output.   However, these terms fail to account for how the information contributed by word $j$ is affected by the LSTM's forget gates between words $j$ and $T$. Consequently, we empirically found that the importance scores from this approach often yield a considerable amount of false positives. A more nuanced approach is obtained by considering the additive decomposition of $c_T$ in equation \eqref{eq:additive}, where each term $e_j$ can be interpreted as the contribution to the cell state $c_T$ by word $j$. By iterating the equation $c_t = f_tc_{t - 1} + i_t \tilde{c}_t$, we get that
\begin{equation}
c_T = \sum_{i=1}^T (\prod_{j=i + 1}^T f_j) i_i \tilde{c}_i = \sum_{i=1}^T e_{i, T} \label{eq:additive}
\end{equation}
This suggests a natural definition of an alternative score to the $\beta_{i, j}$, corresponding to augmenting the $c_j$ terms with products of forget gates to reflect the upstream changes made to $c_j$ after initially processing word $j$. 
\begin{align}
\label{eq:gamma}
\exp(W_i h_T) & = \prod_{j=1}^T \exp \left( W_i(o_T \odot (\tanh(\sum_{k = 1}^j e_{k, T}) - \tanh(\sum_{k=1}^{j - 1}e_{k, T})))\right) \\
& = \prod_{j=1}^T \exp \left( W_i(o_T \odot (\tanh((\prod_{k=j+1}^t f_k) c_j) - \tanh((\prod_{k=j}^t f_k) c_{j - 1}))) \right) \\
& = \prod_{j=1}^T \gamma_{i,j}
\end{align}
\section{Phrase Extraction for Document Classification}
We now introduce a technique for using our variable importance scores to extract phrases from a trained LSTM. To do so, we search for phrases which consistently provide a large contribution to the prediction of a particular class relative to other classes. The utility of these patterns is validated by using them as input for a rules based classifier. For simplicity, we focus on the binary classification case.
\subsection{Phrase Extraction}
\label{sec:phrase_extraction}
A phrase can be reasonably described as predictive if, whenever it occurs, it causes a document to both be labelled as a particular class, and not be labelled as any other. As our importance scores introduced above correspond to the contribution of particular words to class predictions, they can be used to score potential patterns by looking at a pattern's average contribution to the prediction of a given class relative to other classes. More precisely, given a collection of $D$ documents $\{\{x_{i,j}\}_{i = 1}^ {N_d} \}_{j=1}^D$, for a given phrase $w_1,...,w_k$ we can compute scores $S_1, S_2$ for classes 1 and 2, as well as a combined score $S$ and class $C$ as
\begin{align}
S_1(w_1,...,w_k) & =  \frac{\text{Average}_{j,b}\left\lbrace \prod_{l=1}^k \beta_{1, b + l, j} | x_{b + i, j} = w_i, i = 1,...,k  \right\rbrace}{\text{Average}_{j,b}\left\lbrace \prod_{l=1}^k \beta_{2, b + l, j} | x_{b + i, j} = w_i, i = 1,...,k  \right\rbrace} \\
S_2(w_1,..,w_k) & = \frac{1}{S_1(w_1,...,w_k)} \\
S(w_1,...,w_k) & = \max_i(S_i(w_1,...,w_k))\\
C(w_1,...,w_k) & = \text{argmax}_i(S_i(w_1,...,w_k))
\end{align}
where $\beta_{i, j, k}$ denotes $\beta_{i, j}$ applied to document $k$. 

The numerator of $S_1$ denotes the average contribution of the phrase to the prediction of class 1 across all occurrences of the phrase. The denominator denotes the same statistic, but for class 2. Thus, if $S_1$ is high, then $w_1,...,w_k$ is a strong signal for class 1, and likewise for $S_2$. We propose to use $S$ as a score function in order to search for high scoring, representative, phrases which provide insight into the trained LSTM, and $C$ to denote the class corresponding to a phrase.

In practice, the number of phrases is too large to feasibly compute the score of them all. Thus, we approximate a brute force search through a two step procedure. First, we construct a list of candidate phrases by searching for strings of consecutive words $j$ with importance scores $\beta_{i,j} > c$ for any $i$ and some threshold $c$; in the experiments below we use $c=1.1$. Then, we score and rank the set of candidate phrases, which is much smaller than the set of all phrases. 
\subsection{Rules based classifier}
The extracted patterns from Section \ref{sec:phrase_extraction} can be used to construct a simple, rules-based classifier which approximates the output of the original LSTM. Given a document and a list of patterns sorted by descending score given by $S$, the classifier sequentially searches for each pattern within the document using simple string matching. Once it finds a pattern, the classifier returns the associated class given by $C$, ignoring the lower ranked patterns. The resulting classifier is interpretable,  and despite its simplicity,  retains much of the accuracy of the LSTM used to build it.
\section{Experiments}
We now present the results of our experiments. 
\subsection{Training Details}
We implemented all models in Torch using default hyperparameters for weight initializations. For WikiMovies, all documents and questions were pre-processed so that multiple word entities were concatenated into a single word. For a given question, relevant articles were found by first extracting from the question the rarest entity, then returning a list of Wikipedia articles containing any of those words. We use the pre-defined splits into train, validation and test sets, containing 96k, 10k and 10k questions, respectively. The word and hidden representations of the LSTM were both set to dimension 200 for WikiMovies, 300 and 512 for Yelp, and 300 and 150 for Stanford Sentiment Treebank. All models were optimized using Adam \cite{adam} with the default learning rate of 0.001 using early stopping on the validation set. For rule extraction using gradient scores, the product in the reward function is replaced by a sum. In both datasets, we found that normalizing the gradient scores by the largest gradient improved results.
\subsection{Sentiment Analysis}
\label{sec:sentresults}
We first applied the document classification framework to two different sentiment analysis datasets. Originally introduced in \cite{charCNN}, the Yelp review polarity dataset was obtained from the Yelp Dataset Challenge and has train and test sets of size 560,000 and 38,000. The task is binary prediction for whether the review is positive (four or five stars) or negative (one or two stars). The reviews are relatively long, with an average length of 160.1 words. We also used the binary classification task from the Stanford Sentiment Treebank (SST) \cite{SST}, which has less data with train/dev/test sizes of 6920/872/1821, and is done at a sentence level, so has much shorter document lengths.

We report results in Table \ref{sentimentResults} for seven different models. We report state of the art results from prior work  using convolutional neural networks; \cite{KimCNN} for SST and \cite{charCNN} for Yelp. We also report our LSTM baselines, which are competitive with state of the art, along with the three different pattern matching models described above. For SST, we also report prior results using bag of words features with Naive Bayes. 

The additive cell decomposition pattern equals or outperforms the cell-difference patterns, which handily beat the gradient results. This coincides with our empirical observations regarding the information contained within the importance measures, and validates our introduced measure. The differences between measures become more pronounced in Yelp, as the longer document sizes provide more opportunities for false positives. 

Although our pattern matching algorithms underperform other methods, we emphasize that pure performance is not our goal, nor would we expect more from such a simple model. Rather, the fact that our method provides reasonable accuracy is one piece of evidence, in addition to the qualitative evidence given later, that our word importance scores and extracted patterns contain useful information for understanding the actions of a LSTM. 
\begin{center}
\begin{table}
\begin{center}
\begin{tabularx}{.75\textwidth}{@{} Y Y Y @{}}
\hline 
\textbf{Model} & \textbf{Yelp Polarity} &  \textbf{Stanford Sentiment Treebank} \\
\hline
Large word2vec CNN \cite{charCNN}  & 95.4 & - \\
\hline
CNN-multichannel \cite{KimCNN} & - & 88.1 \\
\hline
Naive Bayes \cite{SST} & - & 82.6 \\
\hline
LSTM & 95.3 & 87.3  \\
\hline
Cell Decomposition Pattern Matching & 86.5 & 76.2 \\
\hline
Cell-Difference Pattern Matching & 81.2 & 77.4 \\ 
\hline 
Gradient Pattern Matching & 65.0 & 68.0 \\
\hline
\end{tabularx}
\caption{Test accuracy on sentiment analysis.  See section \ref{sec:sentresults} for further descriptions of the models.}
\label{sentimentResults}
\end{center}
\end{table}
\end{center}
\subsection{WikiMovies}
Although document classification comprises a sizeable portion of current research in natural language processing, much recent work focuses on more complex problems and models. In this section, we examine WikiMovies, a recently introduced question answer dataset, and show that with some simple modifications our approach can be adapted to this problem. 

\subsubsection{Dataset}

WikiMovies is a dataset consisting of more than 100,000 questions about movies, paired with relevant Wikipedia articles. It was constructed using the pre-existing dataset MovieLens, paired with templates extracted from the SimpleQuestions dataset \cite{simpleQuestions}, a open-domain question answering dataset based on Freebase. They then selected a set of Wikipedia articles about movies by identifying a set of movies from OMDb that had an associated article by title match, and kept the title and first section for each article. 

For a given question, the task is to read through the relevant articles and extract the answer, which is contained somewhere within the text. The dataset also provides a list of ~43k entities containing all possible answers.

\subsubsection{LSTMs for WikiMovies}
\label{sec:LSTMforQA}
We propose a simplified version of recent work \cite{baidu}. Given a pair of question $x_1^q,...,x_N^q$ and document $x_1^d,...,x_T^d$, we first compute an embedding for the question using a LSTM. Then, for each word $t$ in the document, we augment the word embedding $x_t$ with the computed question embedding. This is equivalent to adding an additional term which is linear in the question embedding into the gate equations 3-6, allowing the patterns an LSTM absorbs to be directly conditioned upon the question at hand. 
\begin{align}
h_t^q & = \text{LSTM}(x_t^q) \\
h_t & = \text{LSTM}(x_t^d \| h_N^q)
\end{align}
Having run the above model over the document while conditioning on a question, we are given contextual representations $h_1,...,h_T$ of the words in the document. For each entity $t$ in the document we use $p_t$ to conduct a binary prediction for whether or not the entity is the answer. At test time, we return the entity with the highest probability as the answer. 
\begin{equation}
p_t = \text{SoftMax}(W h_t) 
\end{equation}
\subsubsection{Phrase Extraction}
\label{sec:PMforQA}
We now introduce some simple modifications that were useful in adapting our pattern extraction framework to this specific task. First, in order to define the set of classifications problems to search over, we treat each entity $t$ within each document as a separate binary classification task with corresponding predictor $p_t$. Given this set of classification problems, rather than search over the space of all possible phrases, we restrict ourselves to those ending at the entity in question. We also distinguish patterns starting at the beginning of the document with those that do not and introduce an entity character into our pattern vocabulary, which can be matched by any entity. Template examples can be seen below, in Table \ref{learnedPatterns}. Once we have extracted a list of patterns, in the rules-based classifier we only search for positive examples, and return as the answer the entity matched to the highest ranked positive pattern.

\subsubsection{Results}
\label{sec:results}
We report results on six different models in Tables \ref{resultsSummary} and \ref{detResults}. We show the results from \cite{kvMem}, which fit a key-value memory network (KV-MemNN) on representations from information extraction (IE) and raw text (Doc).  Next, we  report the results of the LSTM  described in Section \ref{sec:LSTMforQA}.  Finally, we show the results of using three variants of the pattern matching algorithm described in Section \ref{sec:PMforQA}:  using patterns extracted using the additive decomposition (cell decomposition), difference in cells approaches (cell-difference) and gradient importance scores (gradient), as discussed in Section \ref{sec:relatedWork}. Performance is reported using the accuracy of the top hit over all possible answers (all entities), i.e. the hits@1 metric. 

As shown in Table \ref{resultsSummary}, our LSTM model surpasses the prior state of the art by nearly 4\%. Moreover, our automatic pattern matching model approximates the LSTM with less than 6\% error, which is surprisingly small for such a simple model, and falls within 2\% of the prior state of the art. Similarly to sentiment analysis, we observe a clear ordering of the results across question categories, with our cell decomposition scores providing the best performance, followed by the cell difference and gradient scores. 
\begin{center}
\begin{table}
\begin{center}
\begin{tabularx}{.5\textwidth}{@{} Y Y @{}}
\hline 
\textbf{Model} & \textbf{Test accuracy} \\
\hline
KV-MemNN IE  & 68.3 \\
\hline
KV-MemNN Doc & 76.2 \\
\hline
LSTM & 80.1 \\
\hline
Cell Decomposition Pattern Matching & 74.3 \\
\hline
Cell-Difference Pattern Matching & 69.4 \\ 
\hline 
Gradient Pattern Matching & 57.4 \\
\hline
\end{tabularx}
\caption{Test results on WikiMovies, measured in \% hits@1.  See Section \ref{sec:results} for further descriptions of the models.}
\label{resultsSummary}
\end{center}
\end{table}
\end{center}

\begin{table}
\begin{tabular}{|l|c|c|c|c|c|c|}
\hline 
 & \topbottom{KV-MemNN}{IE} & \topbottom{KV-MemNN}{Doc} & LSTM & \topbottom{Cell Decomp}{RE} & \topbottom{Cell Diff}{RE} & \topbottom{Gradient}{RE}\\
\hline 
Actor to Movie & 66 & 83 & 82 & 78 & 77 & 78 \\ 
\hline 
Director to Movie & 78 & 91 & 84 & 82 & 84 & 83\\ 
\hline 
Writer to Movie & 72 & 91 & 88 & 88 & 89 & 88 \\ 
\hline 
Tag to Movie & 35 & 49 & 49 & 38 & 38 & 38 \\ 
\hline 
Movie to Year & 75 & 89 & 89 & 84 & 84 & 84 \\ 
\hline 
Movie to Writer & 61 & 64 & 86 & 79 & 72 & 63 \\ 
\hline 
Movie to Actor & 64 & 64 & 84 & 75 & 73 & 67 \\ 
\hline 
Movie to Director & 76 & 79 & 88 & 86 & 85 & 45\\ 
\hline 
Movie to Genre & 84 & 86 & 72 & 65 & 42 & 21\\ 
\hline 
Movie to Votes & 92 & 92 & 67 & 67 & 67 & 67\\ 
\hline 
Movie to Rating & 75 & 92 & 33 & 25 & 25 & 25\\ 
\hline 
Movie to Language & 62 & 84 & 72 & 67 & 66 & 44\\ 
\hline 
Movie to Tags &  47 & 48 & 58 & 44 & 30 & 6\\ 
\hline 
\end{tabular} 
\caption{Results broken down by question category.  See section \ref{sec:results} for further descriptions of the models.}
\label{detResults}
\end{table} 

\section{Discussion}

\subsection{Learned patterns}
We present extracted patterns for both sentiment tasks, and some WikiMovies question categories in Table \ref{learnedPatterns}. These patterns are qualitatively sensible, providing further validation of our approach. The increased size of the Yelp dataset allowed for longer phrases to be extracted relative to SST. 

\begin{table}
\begin{tabularx}{\textwidth}{@{} Y Y @{}} 
\toprule
\textbf{Category} & \textbf{Top Patterns} \\
\midrule
\textbf{Yelp Polarity Positive} & definitely come back again., love love love this place, great food and great service., highly recommended!, will definitely be coming back, overall great experience, love everything about, hidden gem. \\
\textbf{Yelp Polarity Negative} & worst customer service ever, horrible horrible horrible, won't be back, disappointed in this place, never go back there, not worth the money, not recommend this place \\
\textbf{SST Positive} & riveting documentary, is a real charmer, funny and touching, well worth your time, journey of the heart, emotional wallop, pleasure to watch, the whole family, cast is uniformly superb, comes from the heart, best films of the year, surprisingly funny, deeply satisfying\\
\textbf{SST Negative} & pretentious mess ..., plain bad, worst film of the year, disappointingly generic, fart jokes, banal dialogue, poorly executed, waste of time, a weak script, dullard, how bad it is, platitudes, never catches fire, tries too hard to be, bad acting, untalented artistes, derivative horror film, lackluster\\ 
\textbf{WikiMovies movie to writer} & film adaptation of Charles Dickens', film adapted from ENT, by journalist ENT, written by ENT\\
\textbf{WikiMovies movie to actor} & western film starring ENT, starring Ben Affleck, . The movie stars ENT, that stars ENT \\
\textbf{WikiMovies movie to language} & is a 2014 french, icelandic, finnish, russian, danish, bengali, dutch, original german, zulu,czech, estonian, mandarin, filipino, hungarian \\
\bottomrule
\end{tabularx}
\caption{Selected top patterns using cell decomposition scores, ENT denotes an entity placeholder}
\label{learnedPatterns}
\end{table}

\subsection{Approximation error between LSTM and pattern matching}
Although our approach is able to extract sensible patterns and achieve reasonable performance, there is still an approximation gap between our algorithm and the LSTM. In Table \ref{errors} we present some examples of instances where the LSTM was able to correctly classify a sentence, and our algorithm was not, along with the pattern used by our algorithm. At first glance, the extracted patterns are sensible, as "gets the job done" or "witty dialogue" are phrases you'd expect to see in a positive review of a movie. However, when placed in the broader context of these particular reviews, they cease to be predictive. This demonstrates that, although our work is useful as a first-order approximation, there are still additional relationships that an LSTM is able to learn from data.  

\begin{table}
\begin{tabularx}{\textwidth}{@{}   p{\dimexpr.15\linewidth-2\tabcolsep-1.3333\arrayrulewidth}
  p{\dimexpr.15\linewidth-2\tabcolsep-1.3333\arrayrulewidth}
  p{\dimexpr.7\linewidth-2\tabcolsep-1.3333\arrayrulewidth}
  @{}}
\toprule
\textbf{Sentiment} & \textbf{Pattern} & \textbf{Sentence} \\
\midrule
Negative & gets the job done & Still, it \hlcyan{gets the job done} --- a sleepy afternoon rental \\
Negative & is a great & This \hlcyan{is a great} subject for a movie, but Hollywood has squandered the opportunity, using is as a prop for a warmed-over melodrama and the kind of choreographed mayhem that director John Woo has built his career on. \\
Negative & happy ending & The story loses its bite in a last-minute \hlcyan{happy ending} that's even less plausible than the rest of the picture.\\
Negative & witty dialogue & An often-deadly boring, strange reading of a classic whose \hlcyan{witty dialogue} is treated with a baffling casual approach. \\
Positive & mess & The film is just a big, gorgeous, mind-blowing, breath-taking \hlcyan{mess}\\
\bottomrule
\end{tabularx}
\caption{Examples from Stanford sentiment treebank which are correctly labelled by our LSTM and incorrectly labelled by our rules-based classifier. The matched pattern is highlighted}
\label{errors}
\end{table}

\subsection{Comparison between word importance measures}
While the prediction accuracy of our rules-based classifier provides quantitative validation of the relative merits of our visualizations, the qualitative differences are also insightful. In Table \ref{ImpComp}, we provide a side-by-side comparison between the different measures. As discussed before, the difference in cells technique fails to account for how the updates resulting from word $j$ are affected by the LSTM's forget gates between when the word is initially processed and the answer. Consequently, we empirically found that without the interluding forget gates to dampen cell movements, the variable importance scores were far noisier than in additive cell decomposition approach. Under the additive cell decomposition, it identifies the phrase 'it stars', as well as the actor's name Aqib Khan as being important, a sensible conclusion. Moreover, the vast majority of words are labelled with an importance score of 1, corresponding to irrelevant. On the other hand, the difference in cells approach yields widely changing importance scores, which are challenging to interpret. In terms of noise, the gradient measures seem to lie somewhere in the middle. These patterns are broadly consistent with what we have observed, and provide qualitative validation of our metrics.

\begin{table}
\begin{tabularx}{\textwidth}{@{} Y Y Y @{}} 
\toprule
\textbf{Additive cell decomposition} & \textbf{Difference in cell values} & \textbf{Gradient} \\
\midrule
\textcolor[RGB]{51, 51, 51}{\fontsize{15.393065929413}{5}\selectfont west is west}
\textcolor[RGB]{51, 51, 51}{\fontsize{15.51902961731}{5}\selectfont is}
\textcolor[RGB]{51, 51, 51}{\fontsize{15.675363063812}{5}\selectfont a}
\textcolor[RGB]{102, 102, 102}{\fontsize{12.629543066025}{5}\selectfont 2010}
\textcolor[RGB]{102, 102, 102}{\fontsize{12.713231563568}{5}\selectfont british comedy}
\textcolor[RGB]{102, 102, 102}{\fontsize{11.966822862625}{5}\selectfont -}
\textcolor[RGB]{102, 102, 102}{\fontsize{12.363042354584}{5}\selectfont drama}
\textcolor[RGB]{102, 102, 102}{\fontsize{11.511328339577}{5}\selectfont film}
\textcolor[RGB]{102, 102, 102}{\fontsize{11.947909712791}{5}\selectfont ,}
\textcolor[RGB]{102, 102, 102}{\fontsize{12.401394367218}{5}\selectfont which}
\textcolor[RGB]{51, 51, 51}{\fontsize{15.579429388046}{5}\selectfont is}
\textcolor[RGB]{102, 102, 102}{\fontsize{13.240346431732}{5}\selectfont a}
\textcolor[RGB]{102, 102, 102}{\fontsize{12.578019618988}{5}\selectfont sequel}
\textcolor[RGB]{102, 102, 102}{\fontsize{12.098379850388}{5}\selectfont to}
\textcolor[RGB]{102, 102, 102}{\fontsize{13.269189119339}{5}\selectfont the}
\textcolor[RGB]{102, 102, 102}{\fontsize{12.495604276657}{5}\selectfont 1999}
\textcolor[RGB]{102, 102, 102}{\fontsize{12.751168727875}{5}\selectfont comedy}
\textcolor[RGB]{102, 102, 102}{\fontsize{11.119393229485}{5}\selectfont "}
\textcolor[RGB]{102, 102, 102}{\fontsize{10.796576857567}{5}\selectfont east is east}
\textcolor[RGB]{51, 51, 51}{\fontsize{16.443211555481}{5}\selectfont "}
\textcolor[RGB]{102, 102, 102}{\fontsize{12.73127746582}{5}\selectfont .}
\textcolor[RGB]{0, 0, 0}{\fontsize{30.765424489975}{5}\selectfont it}
\textcolor[RGB]{51, 51, 51}{\fontsize{17.541554689407}{5}\selectfont stars}
\textcolor[RGB]{0, 0, 0}{\fontsize{29.222506999969}{5}\selectfont aqib khan}

 & 
\textcolor[RGB]{0, 0, 0}{\fontsize{48.989500522614}{5}\selectfont west is west}
\textcolor[RGB]{51, 51, 51}{\fontsize{18.414105892181}{5}\selectfont is}
\textcolor[RGB]{247, 247, 247}{\fontsize{3}{5}\selectfont a}
\textcolor[RGB]{102, 102, 102}{\fontsize{9.73188829422}{5}\selectfont 2010}
\textcolor[RGB]{153, 153, 153}{\fontsize{6.9298274517059}{5}\selectfont british comedy}
\textcolor[RGB]{102, 102, 102}{\fontsize{12.154144763947}{5}\selectfont -}
\textcolor[RGB]{0, 0, 0}{\fontsize{49.131714820862}{5}\selectfont drama}
\textcolor[RGB]{153, 153, 153}{\fontsize{6.5516484975815}{5}\selectfont film}
\textcolor[RGB]{153, 153, 153}{\fontsize{8.2206155061722}{5}\selectfont ,}
\textcolor[RGB]{51, 51, 51}{\fontsize{21.207025766373}{5}\selectfont which}
\textcolor[RGB]{51, 51, 51}{\fontsize{24.787138223648}{5}\selectfont is}
\textcolor[RGB]{153, 153, 153}{\fontsize{3.151567697525}{5}\selectfont a}
\textcolor[RGB]{153, 153, 153}{\fontsize{6.0706819295883}{5}\selectfont sequel}
\textcolor[RGB]{51, 51, 51}{\fontsize{19.400768995285}{5}\selectfont to}
\textcolor[RGB]{102, 102, 102}{\fontsize{9.6258838176727}{5}\selectfont the}
\textcolor[RGB]{102, 102, 102}{\fontsize{10.240971565247}{5}\selectfont 1999}
\textcolor[RGB]{51, 51, 51}{\fontsize{26.068958759308}{5}\selectfont comedy}
\textcolor[RGB]{102, 102, 102}{\fontsize{9.8073692321777}{5}\selectfont "}
\textcolor[RGB]{247, 247, 247}{\fontsize{3}{5}\selectfont east is east}
\textcolor[RGB]{51, 51, 51}{\fontsize{17.626294612885}{5}\selectfont "}
\textcolor[RGB]{153, 153, 153}{\fontsize{4.0424473285675}{5}\selectfont .}
\textcolor[RGB]{0, 0, 0}{\fontsize{27.744606256485}{5}\selectfont it}
\textcolor[RGB]{51, 51, 51}{\fontsize{21.359053373337}{5}\selectfont stars}
\textcolor[RGB]{0, 0, 0}{\fontsize{34.85752415657}{5}\selectfont aqib khan}

& 

\textcolor[RGB]{51, 51, 51}{\fontsize{17.931133031845}{5}\selectfont west is west}
\textcolor[RGB]{51, 51, 51}{\fontsize{15.935043811798}{5}\selectfont is}
\textcolor[RGB]{51, 51, 51}{\fontsize{18.399657726288}{5}\selectfont a}
\textcolor[RGB]{102, 102, 102}{\fontsize{14.663283348083}{5}\selectfont 2010}
\textcolor[RGB]{51, 51, 51}{\fontsize{18.035363674164}{5}\selectfont british comedy}
\textcolor[RGB]{102, 102, 102}{\fontsize{12.51748752594}{5}\selectfont -}
\textcolor[RGB]{102, 102, 102}{\fontsize{12.954791307449}{5}\selectfont drama}
\textcolor[RGB]{102, 102, 102}{\fontsize{13.301186084747}{5}\selectfont film}
\textcolor[RGB]{102, 102, 102}{\fontsize{12.735189914703}{5}\selectfont ,}
\textcolor[RGB]{102, 102, 102}{\fontsize{14.356477975845}{5}\selectfont which}
\textcolor[RGB]{102, 102, 102}{\fontsize{13.774746179581}{5}\selectfont is}
\textcolor[RGB]{51, 51, 51}{\fontsize{18.783095598221}{5}\selectfont a}
\textcolor[RGB]{51, 51, 51}{\fontsize{18.829243898392}{5}\selectfont sequel}
\textcolor[RGB]{51, 51, 51}{\fontsize{21.154944419861}{5}\selectfont to}
\textcolor[RGB]{51, 51, 51}{\fontsize{23.385573148727}{5}\selectfont the}
\textcolor[RGB]{0, 0, 0}{\fontsize{30.211240768433}{5}\selectfont 1999}
\textcolor[RGB]{51, 51, 51}{\fontsize{17.981215238571}{5}\selectfont comedy}
\textcolor[RGB]{51, 51, 51}{\fontsize{19.609430551529}{5}\selectfont "}
\textcolor[RGB]{51, 51, 51}{\fontsize{19.751820087433}{5}\selectfont east is east}
\textcolor[RGB]{51, 51, 51}{\fontsize{23.869829893112}{5}\selectfont "}
\textcolor[RGB]{0, 0, 0}{\fontsize{33.454735994339}{5}\selectfont .}
\textcolor[RGB]{0, 0, 0}{\fontsize{33.404174566269}{5}\selectfont it}
\textcolor[RGB]{0, 0, 0}{\fontsize{35.286255598068}{5}\selectfont stars}
\textcolor[RGB]{0, 0, 0}{\fontsize{37.201939344406}{5}\selectfont aqib khan}
\\
\bottomrule
\end{tabularx}
\caption{Comparison of importance scores acquired by three different approaches, conditioning on the question "the film west is west starred which actors?". Bigger and darker means more important.}
\label{ImpComp}
\end{table}

\section{Conclusion}
In this paper, we introduced a novel method for visualizing the importance of specific inputs in determining the output of an LSTM. By searching for phrases which consistently provide large contributions, we are able to distill trained, state of the art, LSTMs into an ordered set of representative phrases. We quantitatively validate the extracted phrases through their performance in a simple, rules-based classifier. Results are shown in a general document classification framework, then specialized to a more complex, recently introduced, question answer dataset. Our introduced measures provide superior predictive ability and cleaner visualizations relative to prior work. We believe that this represents an exciting new paradigm for analysing the behaviour of LSTM's. 

\section*{Acknowledgements}
This research was partially funded by Air Force grant FA9550-14-1-0016. It was also supported by the Center for Science of Information (CSoI), an US NSF Science and Technology Center, under grant agreement CCF-0939370. 

\bibliography{iclr2017_conference}
\bibliographystyle{iclr2017_conference}

\section{Appendix - Heat Maps}

We provide an example heat map using the cell decomposition metric for each class in both sentiment analysis datasets, and selected WikiMovie question categories

\begin{table}[htb]
\begin{tabularx}{\textwidth}{@{}   p{\dimexpr.1\linewidth-2\tabcolsep-1.3333\arrayrulewidth}
  p{\dimexpr.15\linewidth-2\tabcolsep-1.3333\arrayrulewidth}
  p{\dimexpr.75\linewidth-2\tabcolsep-1.3333\arrayrulewidth}
  @{}}
\toprule
Dataset & Category & Heat Map  \\
\midrule
Yelp Polarity & Positive & \textcolor[RGB]{102, 102, 102}{\fontsize{15.918352603912}{5}\selectfont we}
\textcolor[RGB]{102, 102, 102}{\fontsize{13.86031627655}{5}\selectfont went}
\textcolor[RGB]{102, 102, 102}{\fontsize{15.923917293549}{5}\selectfont here}
\textcolor[RGB]{102, 102, 102}{\fontsize{16.584589481354}{5}\selectfont twice}
\textcolor[RGB]{102, 102, 102}{\fontsize{14.950389862061}{5}\selectfont for}
\textcolor[RGB]{102, 102, 102}{\fontsize{15.224142074585}{5}\selectfont breakfast}
\textcolor[RGB]{102, 102, 102}{\fontsize{15.705654621124}{5}\selectfont .}
\textcolor[RGB]{102, 102, 102}{\fontsize{15.145728588104}{5}\selectfont had}
\textcolor[RGB]{102, 102, 102}{\fontsize{15.070688724518}{5}\selectfont the}
\textcolor[RGB]{51, 51, 51}{\fontsize{18.466494083405}{5}\selectfont bananas}
\textcolor[RGB]{51, 51, 51}{\fontsize{21.34708404541}{5}\selectfont foster}
\textcolor[RGB]{102, 102, 102}{\fontsize{15.904316902161}{5}\selectfont waffles}
\textcolor[RGB]{102, 102, 102}{\fontsize{15.280413627625}{5}\selectfont with}
\textcolor[RGB]{51, 51, 51}{\fontsize{18.252518177032}{5}\selectfont fresh}
\textcolor[RGB]{102, 102, 102}{\fontsize{15.658059120178}{5}\selectfont whipped}
\textcolor[RGB]{102, 102, 102}{\fontsize{15.169723033905}{5}\selectfont cream}
\textcolor[RGB]{102, 102, 102}{\fontsize{14.969524145126}{5}\selectfont ,}
\textcolor[RGB]{102, 102, 102}{\fontsize{15.660099983215}{5}\selectfont they}
\textcolor[RGB]{102, 102, 102}{\fontsize{13.461439609528}{5}\selectfont were}
\textcolor[RGB]{0, 0, 0}{\fontsize{32.636561393738}{5}\selectfont amazing}
\textcolor[RGB]{51, 51, 51}{\fontsize{18.088808059692}{5}\selectfont !}
\textcolor[RGB]{102, 102, 102}{\fontsize{16.053328514099}{5}\selectfont !}
\textcolor[RGB]{0, 0, 0}{\fontsize{56.198573112488}{5}\selectfont perfect}
\textcolor[RGB]{51, 51, 51}{\fontsize{18.857681751251}{5}\selectfont seat}
\textcolor[RGB]{102, 102, 102}{\fontsize{15.860040187836}{5}\selectfont out}
\textcolor[RGB]{102, 102, 102}{\fontsize{16.9855427742}{5}\selectfont side}
\textcolor[RGB]{102, 102, 102}{\fontsize{15.361063480377}{5}\selectfont on}
\textcolor[RGB]{102, 102, 102}{\fontsize{15.15965461731}{5}\selectfont the}
\textcolor[RGB]{51, 51, 51}{\fontsize{17.569072246552}{5}\selectfont terrace}
\\
\midrule 
Yelp Polarity & Negative & \textcolor[RGB]{51, 51, 51}{\fontsize{18.438884492787}{5}\selectfont call}
\textcolor[RGB]{102, 102, 102}{\fontsize{14.848994832963}{5}\selectfont me}
\textcolor[RGB]{0, 0, 0}{\fontsize{27.014783437863}{5}\selectfont spoiled}
\textcolor[RGB]{102, 102, 102}{\fontsize{15.698493928524}{5}\selectfont ...this}
\textcolor[RGB]{102, 102, 102}{\fontsize{15.614690023046}{5}\selectfont sushi}
\textcolor[RGB]{102, 102, 102}{\fontsize{14.814351939554}{5}\selectfont is}
\textcolor[RGB]{0, 0, 0}{\fontsize{43.191443120101}{5}\selectfont gross}
\textcolor[RGB]{51, 51, 51}{\fontsize{22.376131818857}{5}\selectfont and}
\textcolor[RGB]{51, 51, 51}{\fontsize{19.316291205325}{5}\selectfont the}
\textcolor[RGB]{51, 51, 51}{\fontsize{20.156941712805}{5}\selectfont orange}
\textcolor[RGB]{102, 102, 102}{\fontsize{16.69427844533}{5}\selectfont chicken}
\textcolor[RGB]{102, 102, 102}{\fontsize{15.089024232246}{5}\selectfont ,}
\textcolor[RGB]{51, 51, 51}{\fontsize{17.213912175476}{5}\selectfont well}
\textcolor[RGB]{102, 102, 102}{\fontsize{15.288405580044}{5}\selectfont it}
\textcolor[RGB]{102, 102, 102}{\fontsize{15.880151068941}{5}\selectfont was}
\textcolor[RGB]{102, 102, 102}{\fontsize{14.834485788384}{5}\selectfont so}
\textcolor[RGB]{51, 51, 51}{\fontsize{18.017103187307}{5}\selectfont thin}
\textcolor[RGB]{102, 102, 102}{\fontsize{14.788226127553}{5}\selectfont i}
\textcolor[RGB]{51, 51, 51}{\fontsize{17.533221534966}{5}\selectfont don}
\textcolor[RGB]{51, 51, 51}{\fontsize{20.462405645032}{5}\selectfont 't}
\textcolor[RGB]{102, 102, 102}{\fontsize{16.00344129262}{5}\selectfont think}
\textcolor[RGB]{102, 102, 102}{\fontsize{14.322935454372}{5}\selectfont it}
\textcolor[RGB]{102, 102, 102}{\fontsize{14.82811911846}{5}\selectfont had}
\textcolor[RGB]{102, 102, 102}{\fontsize{16.325064241612}{5}\selectfont chicken}
\textcolor[RGB]{102, 102, 102}{\fontsize{15.103334867495}{5}\selectfont in}
\textcolor[RGB]{51, 51, 51}{\fontsize{18.575424120663}{5}\selectfont it.}
\textcolor[RGB]{51, 51, 51}{\fontsize{17.576123631112}{5}\selectfont go}
\textcolor[RGB]{0, 0, 0}{\fontsize{40.080508123899}{5}\selectfont somewhere}
\textcolor[RGB]{51, 51, 51}{\fontsize{23.698612427894}{5}\selectfont else} \\
\midrule
Stanford Sentiment & Positive & \textcolor[RGB]{102, 102, 102}{\fontsize{15.127279758453}{5}\selectfont Whether}
\textcolor[RGB]{102, 102, 102}{\fontsize{14.979684352875}{5}\selectfont or}
\textcolor[RGB]{102, 102, 102}{\fontsize{15.122640132904}{5}\selectfont not}
\textcolor[RGB]{102, 102, 102}{\fontsize{15.36928653717}{5}\selectfont you}
\textcolor[RGB]{102, 102, 102}{\fontsize{15.2498960495}{5}\selectfont 're}
\textcolor[RGB]{102, 102, 102}{\fontsize{15.682487487793}{5}\selectfont enlightened}
\textcolor[RGB]{102, 102, 102}{\fontsize{15.093107223511}{5}\selectfont by}
\textcolor[RGB]{102, 102, 102}{\fontsize{15.168263912201}{5}\selectfont any}
\textcolor[RGB]{102, 102, 102}{\fontsize{15.093359947205}{5}\selectfont of}
\textcolor[RGB]{102, 102, 102}{\fontsize{15.41848897934}{5}\selectfont Derrida}
\textcolor[RGB]{102, 102, 102}{\fontsize{15.271797180176}{5}\selectfont 's}
\textcolor[RGB]{102, 102, 102}{\fontsize{15.034437179565}{5}\selectfont lectures}
\textcolor[RGB]{102, 102, 102}{\fontsize{15.223636627197}{5}\selectfont on}
\textcolor[RGB]{102, 102, 102}{\fontsize{15.174608230591}{5}\selectfont ``}
\textcolor[RGB]{102, 102, 102}{\fontsize{15.15828371048}{5}\selectfont the}
\textcolor[RGB]{102, 102, 102}{\fontsize{15.192177295685}{5}\selectfont other}
\textcolor[RGB]{102, 102, 102}{\fontsize{15.084910392761}{5}\selectfont ''}
\textcolor[RGB]{102, 102, 102}{\fontsize{15.424072742462}{5}\selectfont and}
\textcolor[RGB]{102, 102, 102}{\fontsize{15.20884513855}{5}\selectfont ``}
\textcolor[RGB]{102, 102, 102}{\fontsize{15.309698581696}{5}\selectfont the}
\textcolor[RGB]{102, 102, 102}{\fontsize{15.21312713623}{5}\selectfont self}
\textcolor[RGB]{102, 102, 102}{\fontsize{15.916249752045}{5}\selectfont ,}
\textcolor[RGB]{102, 102, 102}{\fontsize{14.170167446136}{5}\selectfont ''}
\textcolor[RGB]{102, 102, 102}{\fontsize{15.492231845856}{5}\selectfont Derrida}
\textcolor[RGB]{51, 51, 51}{\fontsize{17.069535255432}{5}\selectfont is}
\textcolor[RGB]{51, 51, 51}{\fontsize{17.54292011261}{5}\selectfont an}
\textcolor[RGB]{51, 51, 51}{\fontsize{23.856263160706}{5}\selectfont undeniably}
\textcolor[RGB]{0, 0, 0}{\fontsize{37.14123249054}{5}\selectfont fascinating}
\textcolor[RGB]{51, 51, 51}{\fontsize{23.856914043427}{5}\selectfont and}
\textcolor[RGB]{0, 0, 0}{\fontsize{34.458494186401}{5}\selectfont playful}
\textcolor[RGB]{0, 0, 0}{\fontsize{39.521512985229}{5}\selectfont fellow} \\
\midrule
Stanford Sentiment & Negative & \textcolor[RGB]{102, 102, 102}{\fontsize{15.304849147797}{5}\selectfont ...}
\textcolor[RGB]{102, 102, 102}{\fontsize{16.443469524384}{5}\selectfont begins}
\textcolor[RGB]{102, 102, 102}{\fontsize{14.972122907639}{5}\selectfont with}
\textcolor[RGB]{102, 102, 102}{\fontsize{14.448443651199}{5}\selectfont promise}
\textcolor[RGB]{102, 102, 102}{\fontsize{15.277581214905}{5}\selectfont ,}
\textcolor[RGB]{102, 102, 102}{\fontsize{16.306207180023}{5}\selectfont but}
\textcolor[RGB]{102, 102, 102}{\fontsize{15.439486503601}{5}\selectfont runs}
\textcolor[RGB]{51, 51, 51}{\fontsize{22.161920070648}{5}\selectfont aground}
\textcolor[RGB]{102, 102, 102}{\fontsize{16.168282032013}{5}\selectfont after}
\textcolor[RGB]{102, 102, 102}{\fontsize{16.765961647034}{5}\selectfont being}
\textcolor[RGB]{51, 51, 51}{\fontsize{17.556147575378}{5}\selectfont snared}
\textcolor[RGB]{102, 102, 102}{\fontsize{15.869958400726}{5}\selectfont in}
\textcolor[RGB]{102, 102, 102}{\fontsize{15.95543384552}{5}\selectfont its}
\textcolor[RGB]{51, 51, 51}{\fontsize{18.398804664612}{5}\selectfont own}
\textcolor[RGB]{51, 51, 51}{\fontsize{24.077050685883}{5}\selectfont tangled}
\textcolor[RGB]{0, 0, 0}{\fontsize{27.943148612976}{5}\selectfont plot}
\\
\bottomrule
\end{tabularx}
\end{table}

\begin{table}
\begin{tabularx}{\textwidth}{@{}   p{\dimexpr.1\linewidth-2\tabcolsep-1.3333\arrayrulewidth}
  p{\dimexpr.2\linewidth-2\tabcolsep-1.3333\arrayrulewidth}
  p{\dimexpr.7\linewidth-2\tabcolsep-1.3333\arrayrulewidth}
  @{}}
\toprule
Pattern & Question & Heat Map \\
\midrule
Movie to Year &  What was the release year of another 48 hours? & \textcolor[RGB]{51, 51, 51}{\fontsize{23.18589925766}{5}\selectfont another 48 hrs}
\textcolor[RGB]{0, 0, 0}{\fontsize{28.300569057465}{5}\selectfont is}
\textcolor[RGB]{102, 102, 102}{\fontsize{16.569843292236}{5}\selectfont a}
\textcolor[RGB]{51, 51, 51}{\fontsize{17.68404006958}{5}\selectfont 1990}

 \\
\midrule
Movie to Writer &  Which person wrote the movie last of the dogmen? & 
\textcolor[RGB]{102, 102, 102}{\fontsize{15.040321350098}{5}\selectfont last of the dogmen}
\textcolor[RGB]{51, 51, 51}{\fontsize{19.87254858017}{5}\selectfont is}
\textcolor[RGB]{102, 102, 102}{\fontsize{15.891389846802}{5}\selectfont a}
\textcolor[RGB]{102, 102, 102}{\fontsize{15.3919672966}{5}\selectfont 1995}
\textcolor[RGB]{102, 102, 102}{\fontsize{15.004868507385}{5}\selectfont western}
\textcolor[RGB]{102, 102, 102}{\fontsize{15.089347362518}{5}\selectfont adventure}
\textcolor[RGB]{102, 102, 102}{\fontsize{16.581273078918}{5}\selectfont film}
\textcolor[RGB]{51, 51, 51}{\fontsize{18.88002872467}{5}\selectfont written}
\textcolor[RGB]{102, 102, 102}{\fontsize{15.090062618256}{5}\selectfont and}
\textcolor[RGB]{0, 0, 0}{\fontsize{25.938947200775}{5}\selectfont directed}
\textcolor[RGB]{51, 51, 51}{\fontsize{20.446403026581}{5}\selectfont by}
\textcolor[RGB]{0, 0, 0}{\fontsize{29.308984279633}{5}\selectfont tab murphy}
\\
\midrule
Movie to Actor & Who acted in the movie thunderbolt? & \textcolor[RGB]{102, 102, 102}{\fontsize{14.197118282318}{5}\selectfont thunderbolt}
\textcolor[RGB]{51, 51, 51}{\fontsize{18.469083309174}{5}\selectfont (}
\textcolor[RGB]{102, 102, 102}{\fontsize{14.907529354095}{5}\selectfont )}
\textcolor[RGB]{102, 102, 102}{\fontsize{16.893515586853}{5}\selectfont (}
\textcolor[RGB]{102, 102, 102}{\fontsize{13.835830688477}{5}\selectfont "}
\textcolor[RGB]{102, 102, 102}{\fontsize{14.57844376564}{5}\selectfont piklik}
\textcolor[RGB]{102, 102, 102}{\fontsize{14.723117351532}{5}\selectfont foh}
\textcolor[RGB]{102, 102, 102}{\fontsize{13.441327810287}{5}\selectfont "}
\textcolor[RGB]{102, 102, 102}{\fontsize{13.550491333008}{5}\selectfont )}
\textcolor[RGB]{51, 51, 51}{\fontsize{17.424626350403}{5}\selectfont is}
\textcolor[RGB]{102, 102, 102}{\fontsize{15.413269996643}{5}\selectfont a}
\textcolor[RGB]{102, 102, 102}{\fontsize{15.481913089752}{5}\selectfont 1995}
\textcolor[RGB]{102, 102, 102}{\fontsize{16.038181781769}{5}\selectfont hong kong}
\textcolor[RGB]{102, 102, 102}{\fontsize{14.990319013596}{5}\selectfont action}
\textcolor[RGB]{51, 51, 51}{\fontsize{17.741205692291}{5}\selectfont film}
\textcolor[RGB]{51, 51, 51}{\fontsize{22.867517471313}{5}\selectfont starring}
\textcolor[RGB]{153, 153, 153}{\fontsize{9.9129545688629}{5}\selectfont jackie chan}
\\
\midrule
Movie to Director & Who directed bloody bloody bible camp? & \textcolor[RGB]{102, 102, 102}{\fontsize{14.69761133194}{5}\selectfont bloody bloody bible cam p}
\textcolor[RGB]{51, 51, 51}{\fontsize{18.375895023346}{5}\selectfont is}
\textcolor[RGB]{102, 102, 102}{\fontsize{16.462597846985}{5}\selectfont a}
\textcolor[RGB]{102, 102, 102}{\fontsize{15.092079639435}{5}\selectfont 2012}
\textcolor[RGB]{102, 102, 102}{\fontsize{14.888180494308}{5}\selectfont american}
\textcolor[RGB]{102, 102, 102}{\fontsize{15.265958309174}{5}\selectfont horror}
\textcolor[RGB]{102, 102, 102}{\fontsize{15.104331970215}{5}\selectfont -}
\textcolor[RGB]{102, 102, 102}{\fontsize{15.153160095215}{5}\selectfont comedy}
\textcolor[RGB]{102, 102, 102}{\fontsize{14.735482931137}{5}\selectfont /s}
\textcolor[RGB]{102, 102, 102}{\fontsize{14.795472621918}{5}\selectfont platter}
\textcolor[RGB]{51, 51, 51}{\fontsize{17.219667434692}{5}\selectfont film}
\textcolor[RGB]{102, 102, 102}{\fontsize{15.900964736938}{5}\selectfont .}
\textcolor[RGB]{102, 102, 102}{\fontsize{14.957455396652}{5}\selectfont the}
\textcolor[RGB]{102, 102, 102}{\fontsize{15.211074352264}{5}\selectfont film}
\textcolor[RGB]{102, 102, 102}{\fontsize{14.732140302658}{5}\selectfont was}
\textcolor[RGB]{0, 0, 0}{\fontsize{32.753045558929}{5}\selectfont directed}
\textcolor[RGB]{51, 51, 51}{\fontsize{23.306519985199}{5}\selectfont by}
\textcolor[RGB]{0, 0, 0}{\fontsize{28.720462322235}{5}\selectfont vito trabucco}
\\

\midrule
Movie to Genre &  What genre is trespass in? & \textcolor[RGB]{0, 0, 0}{\fontsize{26.452283859253}{5}\selectfont trespass}
\textcolor[RGB]{0, 0, 0}{\fontsize{34.192624092102}{5}\selectfont is}
\textcolor[RGB]{51, 51, 51}{\fontsize{20.636370182037}{5}\selectfont a}
\textcolor[RGB]{153, 153, 153}{\fontsize{12.796659469604}{5}\selectfont 1992}
\textcolor[RGB]{153, 153, 153}{\fontsize{6.4761424064636}{5}\selectfont action}
\\
\midrule
Movie to Votes & How would people rate the pool?  & \textcolor[RGB]{102, 102, 102}{\fontsize{16.849629878998}{5}\selectfont though}
\textcolor[RGB]{102, 102, 102}{\fontsize{14.893523454666}{5}\selectfont filmed}
\textcolor[RGB]{102, 102, 102}{\fontsize{14.237483739853}{5}\selectfont in}
\textcolor[RGB]{153, 153, 153}{\fontsize{12.812790870667}{5}\selectfont hindi}
\textcolor[RGB]{102, 102, 102}{\fontsize{14.32657957077}{5}\selectfont ,}
\textcolor[RGB]{102, 102, 102}{\fontsize{14.522277116776}{5}\selectfont a}
\textcolor[RGB]{51, 51, 51}{\fontsize{18.380868434906}{5}\selectfont language}
\textcolor[RGB]{102, 102, 102}{\fontsize{13.908357620239}{5}\selectfont smith}
\textcolor[RGB]{153, 153, 153}{\fontsize{6.9293713569641}{5}\selectfont didn}
\textcolor[RGB]{51, 51, 51}{\fontsize{20.088863372803}{5}\selectfont 't}
\textcolor[RGB]{102, 102, 102}{\fontsize{16.309413909912}{5}\selectfont know}
\textcolor[RGB]{153, 153, 153}{\fontsize{10.075327157974}{5}\selectfont ,}
\textcolor[RGB]{51, 51, 51}{\fontsize{21.208038330078}{5}\selectfont the}
\textcolor[RGB]{102, 102, 102}{\fontsize{15.882177352905}{5}\selectfont film}
\textcolor[RGB]{102, 102, 102}{\fontsize{16.934356689453}{5}\selectfont earned}
\textcolor[RGB]{0, 0, 0}{\fontsize{60.56182861328}{5}\selectfont $\text{good}^*$}

\\
\midrule
Movie to Rating &  How popular was les miserables?  & \textcolor[RGB]{0, 0, 0}{\fontsize{45.03936290741}{5}\selectfont les mis rables}
\textcolor[RGB]{102, 102, 102}{\fontsize{16.685492992401}{5}\selectfont is}
\textcolor[RGB]{51, 51, 51}{\fontsize{17.432172298431}{5}\selectfont a}
\textcolor[RGB]{102, 102, 102}{\fontsize{16.715741157532}{5}\selectfont 1935}
\textcolor[RGB]{102, 102, 102}{\fontsize{16.392815113068}{5}\selectfont american}
\textcolor[RGB]{102, 102, 102}{\fontsize{15.863442420959}{5}\selectfont drama}
\textcolor[RGB]{51, 51, 51}{\fontsize{20.409224033356}{5}\selectfont film}
\textcolor[RGB]{102, 102, 102}{\fontsize{15.442113876343}{5}\selectfont starring}
\textcolor[RGB]{51, 51, 51}{\fontsize{17.185742855072}{5}\selectfont fredric march}
\textcolor[RGB]{51, 51, 51}{\fontsize{19.489691257477}{5}\selectfont and}
\textcolor[RGB]{51, 51, 51}{\fontsize{19.513516426086}{5}\selectfont charles laughton}
\textcolor[RGB]{51, 51, 51}{\fontsize{20.248744487762}{5}\selectfont based}
\textcolor[RGB]{102, 102, 102}{\fontsize{14.456961154938}{5}\selectfont upon}
\textcolor[RGB]{102, 102, 102}{\fontsize{13.239349126816}{5}\selectfont the}
\textcolor[RGB]{0, 0, 0}{\fontsize{40.918641090393}{5}\selectfont famous}
 \\
\midrule
Movie to Tags &  Describe rough magic? &\textcolor[RGB]{102, 102, 102}{\fontsize{16.084606647491}{5}\selectfont rough magic}
\textcolor[RGB]{51, 51, 51}{\fontsize{18.339736461639}{5}\selectfont is}
\textcolor[RGB]{102, 102, 102}{\fontsize{15.943524837494}{5}\selectfont a}
\textcolor[RGB]{102, 102, 102}{\fontsize{15.428647994995}{5}\selectfont 1995}
\textcolor[RGB]{102, 102, 102}{\fontsize{14.617514610291}{5}\selectfont comedy}
\textcolor[RGB]{102, 102, 102}{\fontsize{16.470649242401}{5}\selectfont film}
\textcolor[RGB]{51, 51, 51}{\fontsize{23.080341815948}{5}\selectfont directed}
\textcolor[RGB]{51, 51, 51}{\fontsize{17.741792201996}{5}\selectfont by}
\textcolor[RGB]{153, 153, 153}{\fontsize{12.850861549377}{5}\selectfont clare peploe}
\textcolor[RGB]{153, 153, 153}{\fontsize{11.716047525406}{5}\selectfont and}
\textcolor[RGB]{51, 51, 51}{\fontsize{18.950564861298}{5}\selectfont starring}
\textcolor[RGB]{102, 102, 102}{\fontsize{14.44725394249}{5}\selectfont bridget fonda}
\textcolor[RGB]{153, 153, 153}{\fontsize{10.830730199814}{5}\selectfont ,}
\textcolor[RGB]{0, 0, 0}{\fontsize{25.346186161041}{5}\selectfont russell crowe}
\\
\midrule
Movie to Language &  What is the main language in fate? & \textcolor[RGB]{0, 0, 0}{\fontsize{32.921435832977}{5}\selectfont fate}
\textcolor[RGB]{51, 51, 51}{\fontsize{21.298835277557}{5}\selectfont (}
\textcolor[RGB]{153, 153, 153}{\fontsize{10.233192443848}{5}\selectfont )}
\textcolor[RGB]{51, 51, 51}{\fontsize{17.003877162933}{5}\selectfont is}
\textcolor[RGB]{102, 102, 102}{\fontsize{14.160645008087}{5}\selectfont a}
\textcolor[RGB]{102, 102, 102}{\fontsize{13.522553443909}{5}\selectfont 2001}
\textcolor[RGB]{51, 51, 51}{\fontsize{19.048819541931}{5}\selectfont turkish}

\\
\bottomrule
\end{tabularx}
\end{table}

\end{document}